\documentclass[conference]{IEEEtran}
\IEEEoverridecommandlockouts
\usepackage{graphicx}
\usepackage{hyperref}
\usepackage[capitalize]{cleveref}
\usepackage{subcaption}
\usepackage{tikz} 
\usepackage{eso-pic}
\newcommand{\md} {\textsc{MCTS-dvc}}
\newcommand{\cpu} {\textsc{MCTS-cpu}}
\newcommand{\gpu} {\textsc{MCTS-gpu}}
\hypersetup{hidelinks}
\usepackage{xcolor}
\def\BibTeX{{\rm B\kern-.05em{\sc i\kern-.025em b}\kern-.08em
    T\kern-.1667em\lower.7ex\hbox{E}\kern-.125emX}}

\begin{document}
\title{Development and Application of a Monte Carlo Tree Search Algorithm for Simulating Da Vinci Code Game Strategies}

\author{Ye Zhang$^{1,*}$, Mengran Zhu$^2$, Kailin Gui$^3$, Jiayue Yu$^4$, Yong Hao$^5$, Haozhan Sun$^6$
\thanks{$^{1,*}$Ye Zhang be with University of Pittsburgh at Pittsburgh, PA 15213, USA {\tt\small \{yez12\}@pitt.edu}}
\thanks{$^2$Mengran Zhu be with Miami University at Oxford, OH 45056, USA {\tt\small \{mengran.zhu0504\}@gmail.com}}
\thanks{$^3$Kailin Gui be with University of Washington at Seattle, WA 98195, USA {\tt\small \{guikailin015\}@gmail.com}}
\thanks{$^4$Jiayue Yu be with Warner Bro. Discovery at Culver City, CA 90232, USA {\tt\small \{jiy048\}@gmail.com}}
\thanks{$^5$Yong Hao be with Columbia University at New York, NY 10027, USA {\tt\small \{EricHao3290\}@gmail.com}}
\thanks{$^6$Haozhan Sun be with Duke University at Durham, NC 27708, USA {\tt\small \{yzjshz1998\}@outlook.com}}
}

\maketitle

\begin{abstract}
In this study, we explore the efficiency of the Monte Carlo Tree Search (MCTS), a prominent decision-making algorithm renowned for its effectiveness in complex decision environments, contingent upon the volume of simulations conducted. Notwithstanding its broad applicability, the algorithm's performance can be adversely impacted in certain scenarios, particularly within the domain of game strategy development. This research posits that the inherent branch divergence within the Da Vinci Code board game significantly impedes parallelism when executed on Graphics Processing Units (GPUs). To investigate this hypothesis, we implemented and meticulously evaluated two variants of the MCTS algorithm, specifically designed to assess the impact of branch divergence on computational performance. Our comparative analysis reveals a linear improvement in performance with the CPU-based implementation, in stark contrast to the GPU implementation, which exhibits a non-linear enhancement pattern and discernible performance troughs. These findings contribute to a deeper understanding of the MCTS algorithm's behavior in divergent branch scenarios, highlighting critical considerations for optimizing game strategy algorithms on parallel computing architectures.
\end{abstract}



\section{Introduction}
\setlength{\parskip}{0.1em}
The landmark achievement of artificial intelligence surpassing human proficiency in the game of Go marked a pivotal moment in the field. Traditionally regarded as an insurmountable challenge due to its vast complexity, far exceeding that of chess or Korean chess, Go was a benchmark for AI capabilities. The victory of AlphaGo over Sedol Lee, a top Korean Go player, followed by its success against Ke Jie, the world's leading player at the time, underscored this breakthrough. A particularly notable aspect of AlphaGo's play was its unconventional move placement, which defied the understanding of seasoned Go players, sparking widespread interest and study among the Go community to decipher these novel strategies~\cite{xu20193d,shi2019data,hu2019real,yan2024self}.

The underlying technology enabling AlphaGo's innovative gameplay is the Monte Carlo Tree Search (MCTS), a decision-making algorithm that leverages extensive simulations to predict outcomes~\cite{silver2016mastering_alphago,zou2022unified}. This algorithm's effectiveness is proportional to the number of simulations it performs, yet the time constraints of Go necessitate a balance between thoroughness and feasibility within each turn's limited duration.
 
To overcome these constraints, AlphaGo was equipped with substantial computational resources, utilizing 48 CPUs and 8 GPUs for parallel processing of simulation cases. This combination of algorithmic refinement and hardware enhancement was crucial for achieving superhuman performance. Nonetheless, the performance of MCTS, especially when executed on GPUs, can be adversely affected by specific characteristics of game algorithms, such as branch divergence~\cite{weimin2024enhancing}. This phenomenon, inherent to the Single Instruction, Multiple Thread (SIMT) architecture, diminishes parallel efficiency due to the need to process both outcomes of conditional branches.

This study posits that the board game 'Da Vinci Code' exemplifies a scenario where branch divergence significantly impacts decision-making efficacy\cite{zhang2020manipulator}. This impact manifests in two primary ways: reduced SIMD utilization due to variable execution path lengths across threads, and diminished simulation capacity per case within fixed time frames, attributed to an increase in the number of MCTS scenarios. The gameplay mechanic of guessing opponents' tiles, with successful guesses allowing additional moves, inherently contributes to branch divergence~\cite{zou2017labeled}.

Further discussion on the unique aspects of `Da Vinci Code' and its interaction with MCTS is provided in later sections, detailing the game's rules (Section~\ref{sec:davinci}) and its implications for MCTS implementation (Section~\ref{sec:mcts}).

Our investigation involved the development and assessment of two MCTS variants to evaluate the impact of branch divergence on GPU performance. We created a CPU-based version using the OpenMP framework~\cite{10405429} for parallel simulation, and a GPU-based version utilizing the CUDA library. The comparative analysis~\cite{wang2024jointly} revealed distinct performance patterns: linear improvement with increased threading on CPU due to isolated core execution, contrasted with non-linear gains on GPU where thread groups share cores (SMs), highlighting scalability issues and performance degradation linked to branch divergence and memory contention.

\section{Background}

\begin{figure}
\begin{subfigure}[b]{0.95\columnwidth}
\includegraphics[width=0.95\columnwidth]{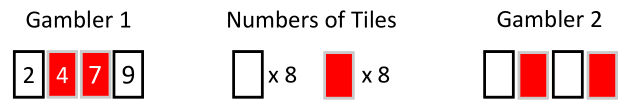}
\caption{Initial setting}
\label{fig:DVC_procedure_1}
\end{subfigure}
\par\smallskip
\begin{subfigure}[b]{0.95\columnwidth}
\includegraphics[width=0.95\columnwidth]{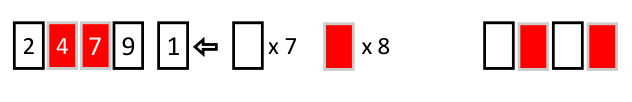}
\caption{Tile selection of palyer 1 (turn 1)}
\label{fig:DVC_procedure_2}
\end{subfigure}
\par\smallskip
\begin{subfigure}[b]{0.95\columnwidth}
\includegraphics[width=0.95\columnwidth]{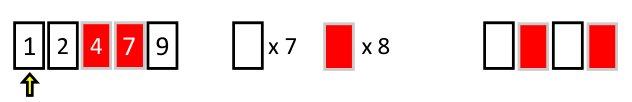}
\caption{Tile insertion of gambler~1 (turn 1)}
\label{fig:DVC_procedure_3}
\end{subfigure}
\par\smallskip
\begin{subfigure}[b]{0.95\columnwidth}
\includegraphics[width=0.95\columnwidth]{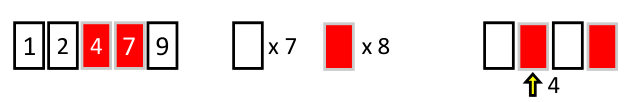}
\caption{Tile guessing of gambler~1 (turn 1)}
\label{fig:DVC_procedure_4}
\end{subfigure}
\par\smallskip
\begin{subfigure}[b]{0.95\columnwidth}
\includegraphics[width=0.95\columnwidth]{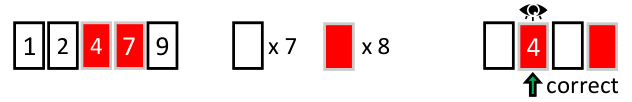}
\caption{A case that the gussing is correct (turn 1)}
\label{fig:DVC_procedure_5}
\end{subfigure}
\par\smallskip
\begin{subfigure}[b]{0.95\columnwidth}
\includegraphics[width=0.95\columnwidth]{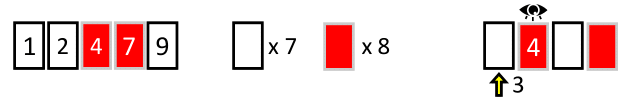}
\caption{Consecutive tile guessing of gambler~1 (turn 1)}
\label{fig:DVC_procedure_6}
\end{subfigure}
\par\smallskip
\begin{subfigure}[b]{0.95\columnwidth}
\includegraphics[width=0.95\columnwidth]{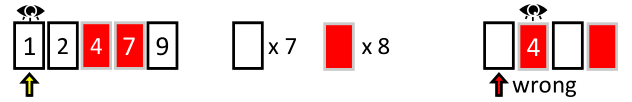}
\caption{A case that the guessing is wrong (turn 1)}
\label{fig:DVC_procedure_7}
\end{subfigure}
\par\smallskip
\begin{subfigure}[b]{0.95\columnwidth}
\includegraphics[width=0.95\columnwidth]{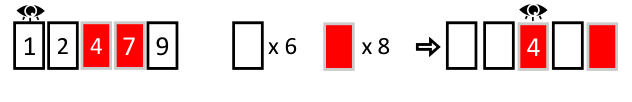}
\caption{Tile selection of gambler~2 (turn 2)}
\label{fig:DVC_procedure_8}
\end{subfigure}
\par\smallskip
\begin{subfigure}[b]{0.95\columnwidth}
\includegraphics[width=0.95\columnwidth]{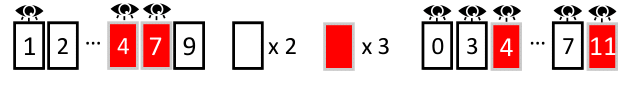}
\caption{Winning state of gambler~1 (turn n)}
\label{fig:DVC_procedure_9}
\end{subfigure}
\caption{Play example of Da Vinci Code}
\end{figure}

\subsection{Da Vinci Code} \label{sec:davinci}

The Da Vinci Code is a strategic board game where players endeavor to deduce the identities of their opponents' tiles. The game comprises 26 tiles, including two jokers and 24 numbered tiles ranging from 0 to 11, evenly divided between black and white colors. This assortment results in a set of black tiles numbered 0 to 11, one black joker, an equivalent set of white tiles, and one white joker. Victory is achieved by the last gambler standing after correctly guessing all the tiles of their opponents.

For illustrative purposes, we delineate the game's mechanics with a structured example\cite{li2024research}. Initially, tiles are obscured on the table, preventing any gambler from viewing their values. Each participant selects four tiles from this pool, organizing them in ascending order from left to right, ensuring the sequence remains visible only to the tile's owner (Figure~\ref{fig:DVC_procedure_1}).

A gambler's turn is bifurcated into two primary actions. The initial phase involves drawing a tile from the remaining pool (Figure~\ref{fig:DVC_procedure_2}) and placing it in its correct position within their own sequence, adhering to the ascending order rule (Figure~\ref{fig:DVC_procedure_3}). Following this, the gambler attempts to guess the value of an opponent's tile. For instance, should gambler 1 posit that gambler 2's first black tile is a 4 (Figure~\ref{fig:DVC_procedure_4}), and this guess proves accurate, gambler 2 is compelled to reveal the guessed tile (Figure~\ref{fig:DVC_procedure_5}). A successful guess grants the guessing gambler an additional attempt. This process is demonstrated when the gambler elects to make another guess (Figure~\ref{fig:DVC_procedure_6}). Conversely, an incorrect guess necessitates the revelation of the newly added tile by the guesser (Figure~\ref{fig:DVC_procedure_7}). The turn concludes either upon a wrong guess or the gambler's decision to cease guessing, thereafter transitioning to the next gambler who then engages in the same two-phased turn (Figure~\ref{fig:DVC_procedure_8}). The game concludes when one gambler successfully unveils all tiles belonging to their opponents, thereby securing a win, as illustrated when gambler 1 reveals all of gambler 2's tiles (Figure~\ref{fig:DVC_procedure_9}).

This explicit breakdown of the Da Vinci Code game not only clarifies its rules and gameplay but also sets the stage for discussing its computational modeling, particularly in relation to the Monte Carlo Tree Search algorithm's handling of decision-making complexities introduced by the game's mechanics.

\subsection{Monte Carlo Tree Search} \label{sec:mcts}

MCTS is a heuristic search algorithm for making a decision.The focus of MCTS is on the analysis of the most successful moves, expanding the search tree based on random sampling of the search space.
The MCTS in games is based on many cases of simulations (or playout)\cite{Tian_Xiang_Feng_Yang_Liu_2024}.
In each simulation, the game is played out by selecting decision at random\cite{zou2023multidimensional}.
The final game result of each simulation is used to weight the nodes in the game tree so that better nodes are more likely to be chosen in future simulations.

The Monte Carlo Tree Search (MCTS) algorithm is pivotal in decision-making, especially in game theory, striking a balance between exploring new strategies and exploiting known ones by constructing a search tree through random sampling\cite{zang2024evaluating,zang2024precision}. It utilizes game simulations to prioritize nodes within the tree, guiding decisions towards more favorable outcomes. This iterative optimization process enables MCTS to navigate complex decision spaces effectively, enhancing the strategic analysis of gameplay\cite{jiang2022deep}.


\section{Method}

\begin{figure}
\begin{subfigure}[b]{0.95\columnwidth}
\includegraphics [width=0.8\columnwidth]{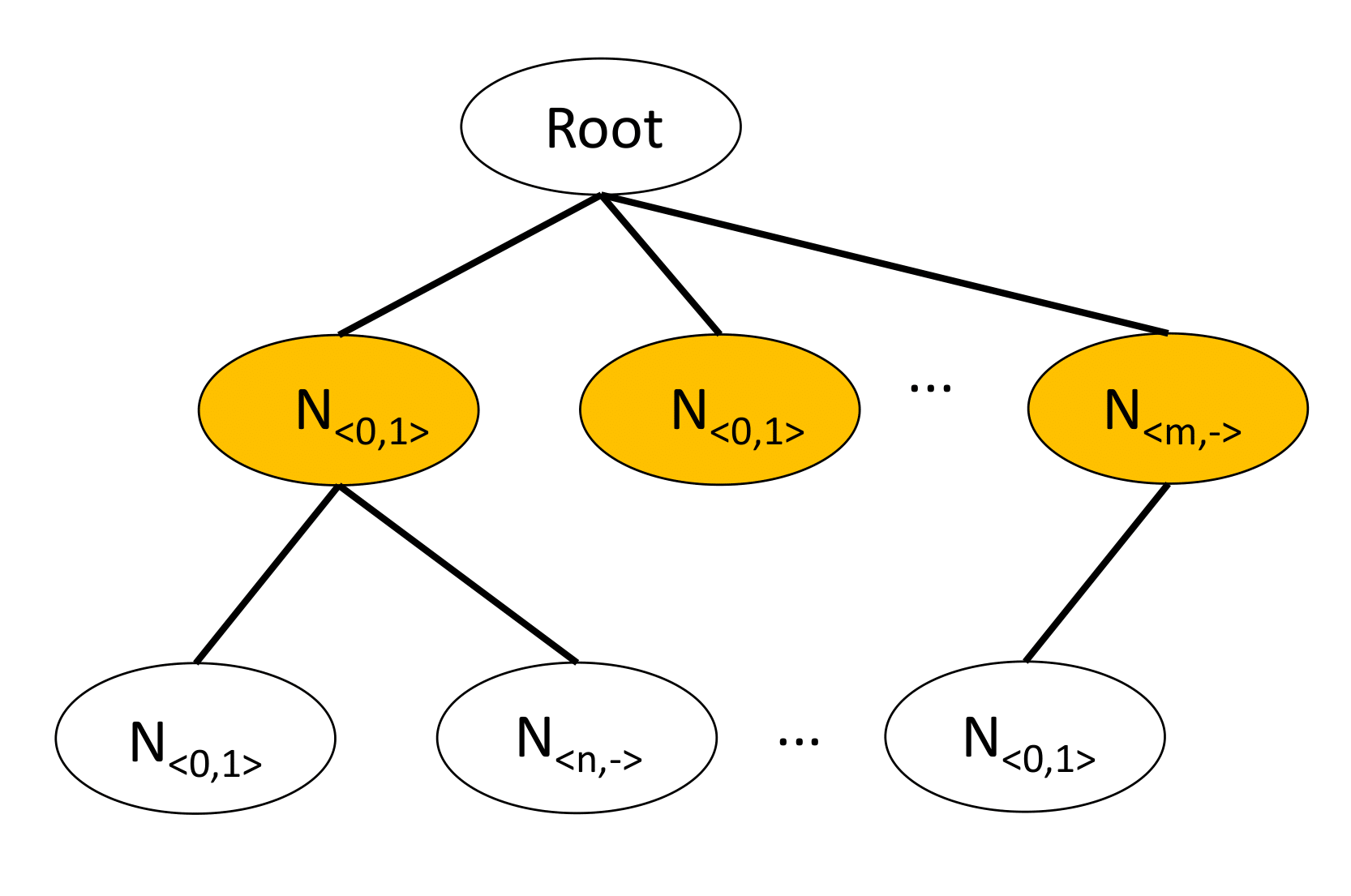}
\caption{MCST with deciding whether to proceed with the game }
\label{fig:proceed_game}
\end{subfigure}
\par\bigskip
\begin{subfigure}[b]{0.95\columnwidth}
\includegraphics [width=0.8\columnwidth]{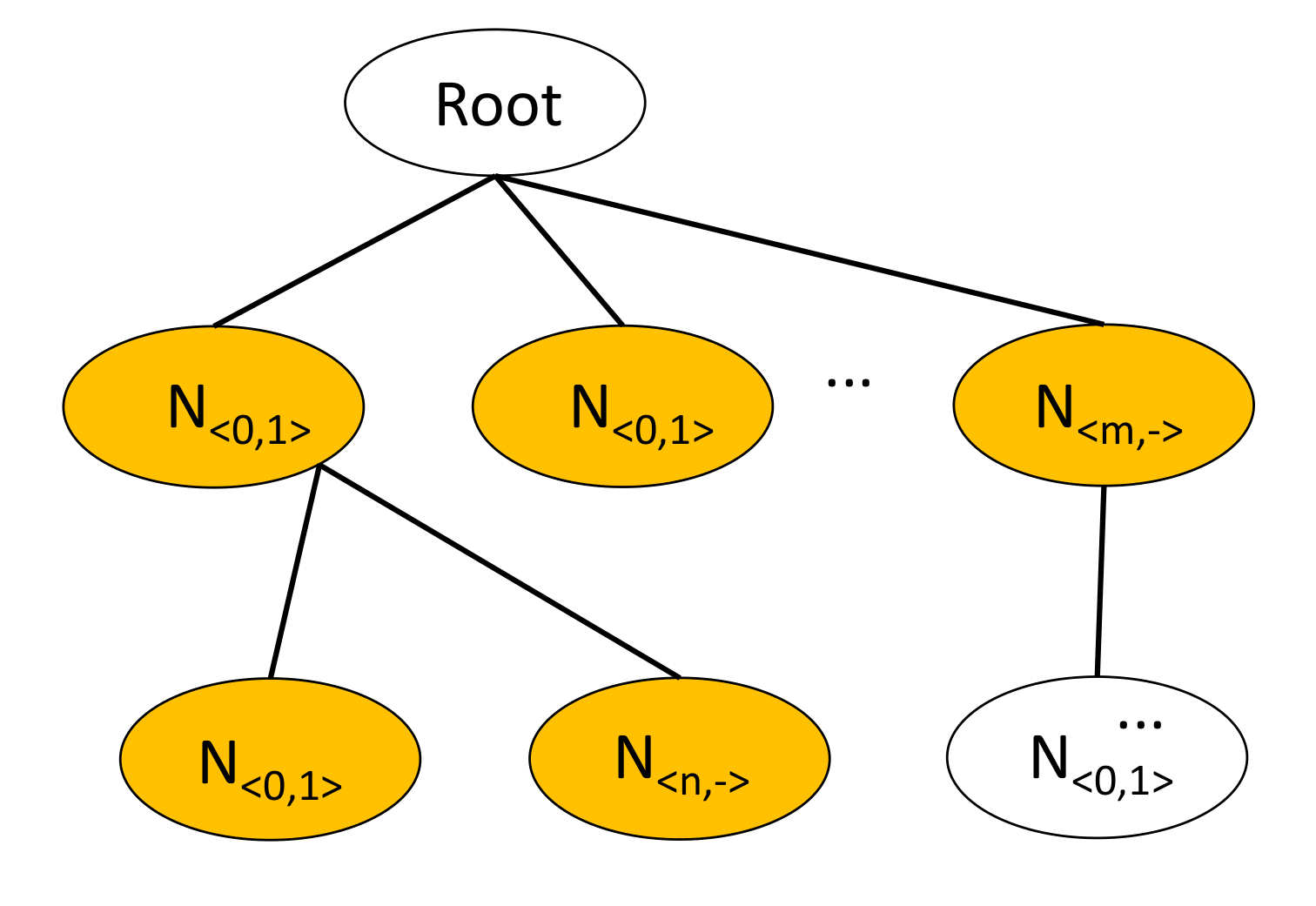}
\caption{MCST with only stop choice}
\label{fig:stop_game}
\end{subfigure}
\caption{Simplification: not to keep guessing}
\end{figure}

In this study, we developed dual variants of the Monte Carlo Tree Search (MCTS) algorithm tailored for the Da Vinci Code game, differentiated by their computational environments. The first variant ($\cpu$) operates within a CPU-based framework, leveraging the OpenMP library for parallelization. Conversely, the second variant ($\gpu$) is designed for a GPU-based environment, utilizing CUDA for both implementation and parallel execution.

Subsequent sections provide an in-depth analysis of each algorithmic variant. Modifications from the standard MCTS approach were necessitated by computational constraints, prompting a detailed exploration of the design considerations that distinguish our adapted models from the conventional MCTS algorithm.

\subsection{Designing Simplified Algorithm}

In our approach, the tree structure is designed such that each node encapsulates a player's guess, characterized by the tile's position and the speculated number. The depth of the tree correlates with the sequence of guesses leading to a particular node, reflecting the iterative nature of decision-making in the Da Vinci Code game. Given players' inherent uncertainty regarding their opponents' tile numbers, our simulations operate under a black-box model. We address this challenge by estimating plausible numbers for the opponent's tiles\cite{song2023energy}. A set of these plausible numbers is randomly selected for each simulation to guide the gameplay, resulting in varied outcomes from identical guesses across different simulations. To accommodate this variability, nodes are enriched with information regarding the chosen set of numbers, allowing for distinct differentiation of outcomes\cite{wei2024narrowing}. This methodology underpins our adaptation of the Monte Carlo Tree Search algorithm for the Da Vinci Code game, referred to as the vanilla Da Vinci Code algorithm (\md).

In refining the \md~algorithm for the Da Vinci Code game, we implemented three key modifications to streamline the decision-making process~\cite{song2023analysis}. A significant adjustment involved disregarding the consideration of plausible numbers for each node. Initially, factoring in these likely numbers resulted in an exponential increase in the potential child nodes, with the root node alone potentially generating up to 11,880 child nodes based on a conservative estimate where the opponent holds four tiles of the same color. This figure starkly contrasts with the 361 initial moves in Go, highlighting the computational inefficiency inherent in managing such a vast decision space\cite{wang2024sub2full}. By omitting the plausible number sets and focusing solely on the position and guessed number, we significantly reduced the complexity of the tree, narrowing down the potential cases to a more manageable 88, thereby enhancing the decision-making quality by mitigating the dilution effect caused by an overly expansive tree structure\cite{wang2022compressive}.

\begin{figure}
\includegraphics[width=1\columnwidth]{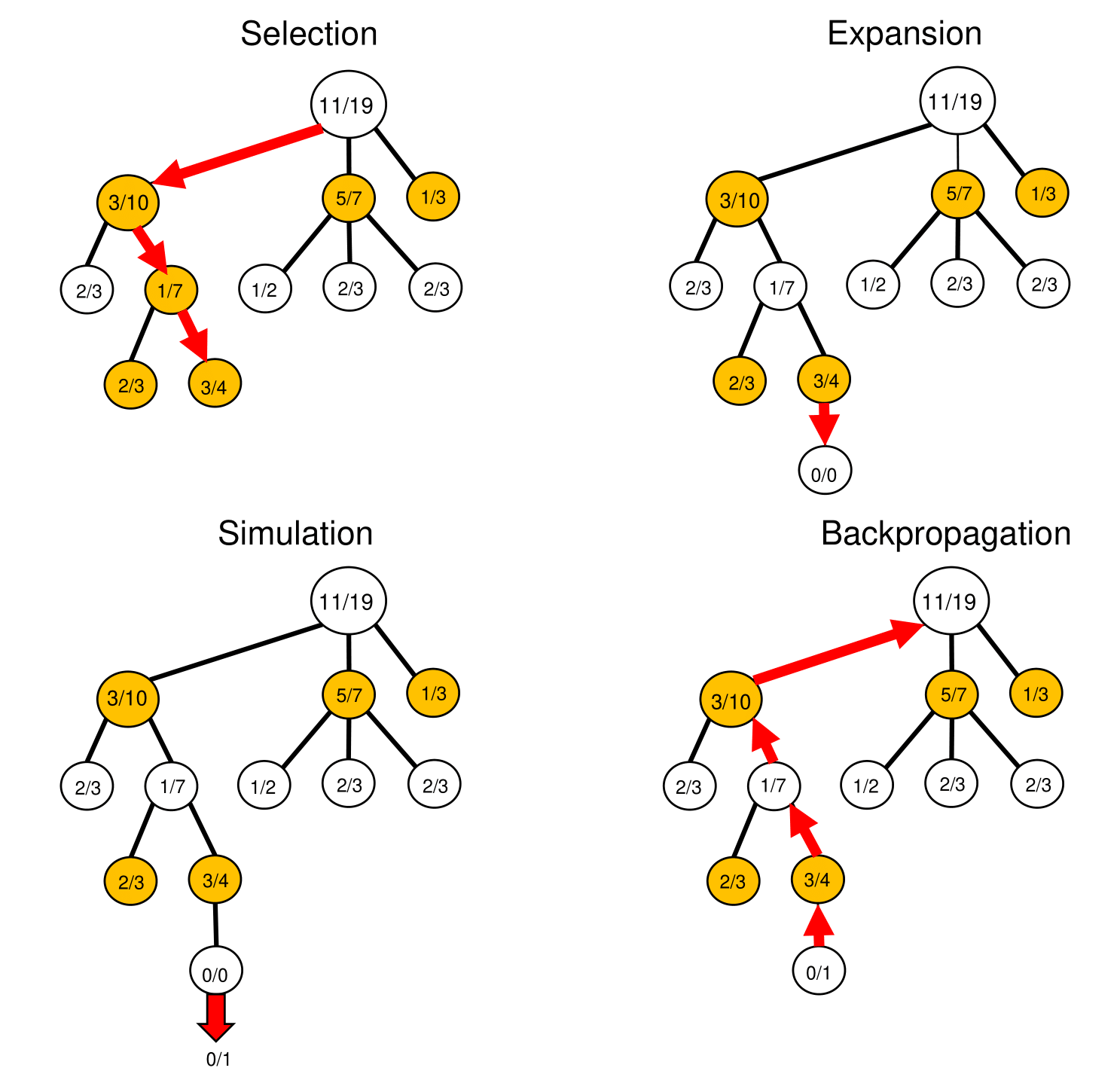}
\caption{Four steps of MCTS}
\label{fig:MCTS_step}
\end{figure}

In our second modification to the \md~algorithm, we addressed the complexity introduced by the option of consecutive guessing. In the original game mechanics, a player successful in guessing a tile could opt to continue guessing or stop. This flexibility results in the potential for varying player nodes at the same depth within the decision tree, as depicted in designated figures, with gray nodes indicating continued guessing by the same player. Such a mechanism significantly complicates the tree structure. To streamline the process and reduce complexity, we altered the Da Vinci Code game rules within our model to mandate cessation of guessing after a correct guess. This simplification not only facilitated a more manageable decision tree but also enabled the efficient implementation of the algorithm\cite{zou2023joint} on both CPU ($\cpu$) and GPU ($\gpu$) platforms by eliminating the need to account for an extensive and complex set of child nodes.
 
\begin{figure}
\begin{subfigure}[b]{0.95\columnwidth}
\includegraphics [width=0.8\columnwidth]{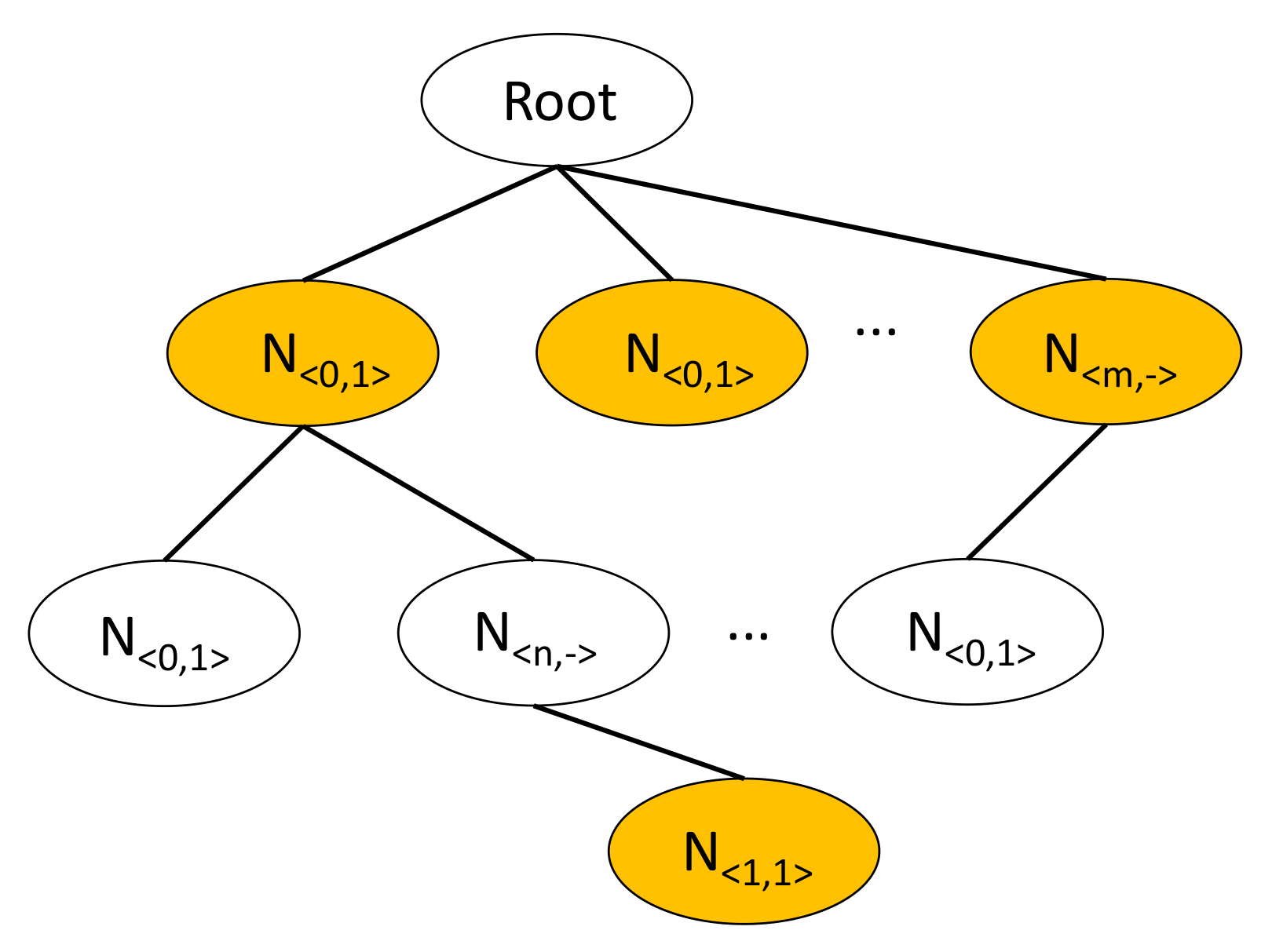}
\caption{MCST with unlimited expansion}
\label{fig:expansion}
\end{subfigure}
\par\bigskip
\begin{subfigure}[b]{0.95\columnwidth}
\includegraphics [width=0.8\columnwidth]{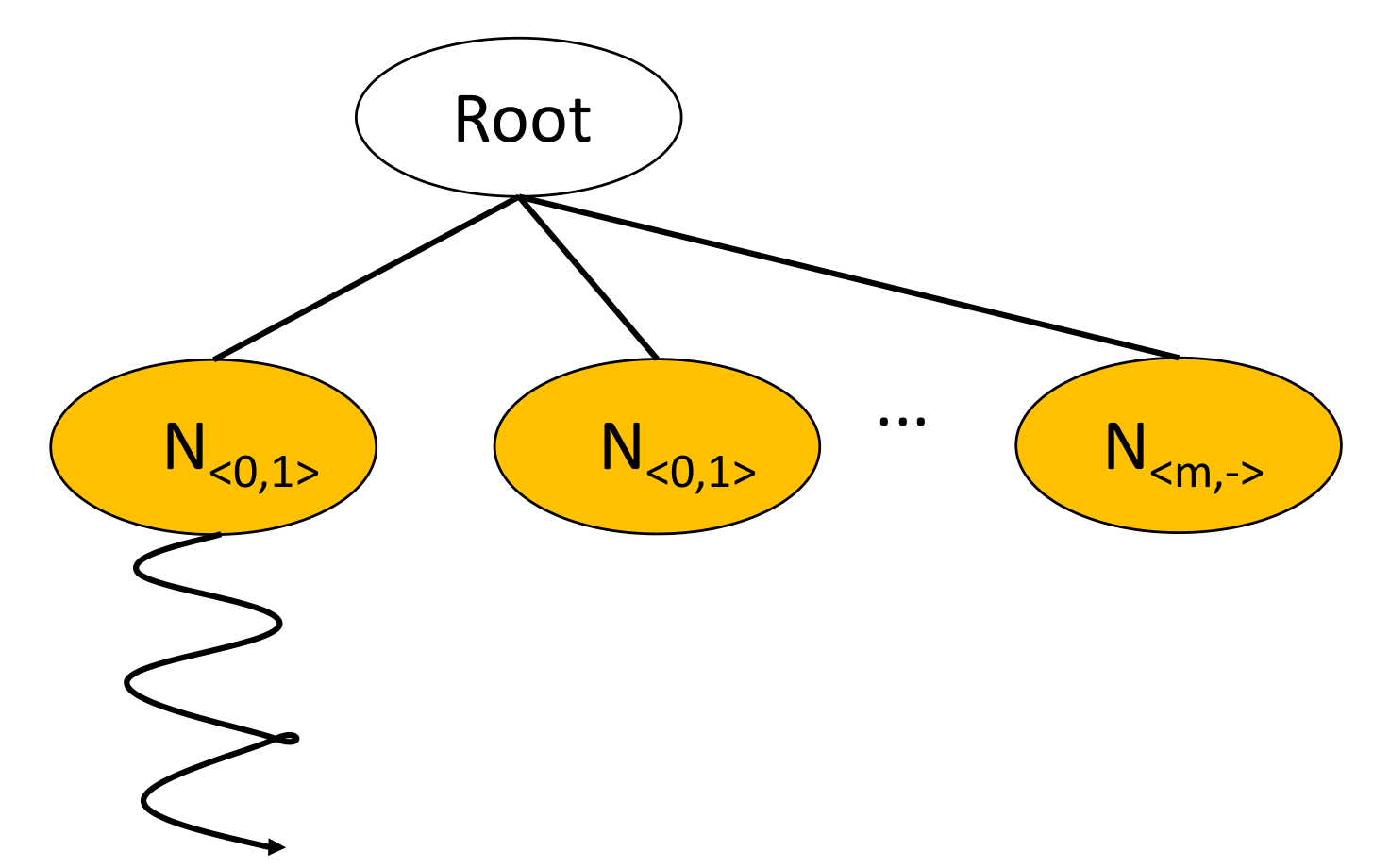}
\caption{MCST with limited expansion}
\label{fig:limited_expansion}
\end{subfigure}
\caption{Simplification: limited expansion}
\end{figure}

Finally, to further refine the \md~algorithm, we imposed a limit on the tree's expansion depth. Given the potential for an infinite number of guesses in the Da Vinci Code game when predictions fail, the theoretical depth\cite{wang2023axial} of the decision tree could be boundless, negatively impacting the algorithm's efficiency. To mitigate this, we established a fixed depth threshold, beyond which nodes engage solely in simulation and backpropagation without further expansion. Figures \cref{fig:expansion} and \cref{fig:limited_expansion} illustrate the distinction between the traditional MCTS approach and our adapted methodology. This strategic limitation of expansion depth effectively precludes nodes of impractical depth, optimizing the algorithm's performance.

\begin{figure}
\includegraphics[width=0.95\columnwidth]{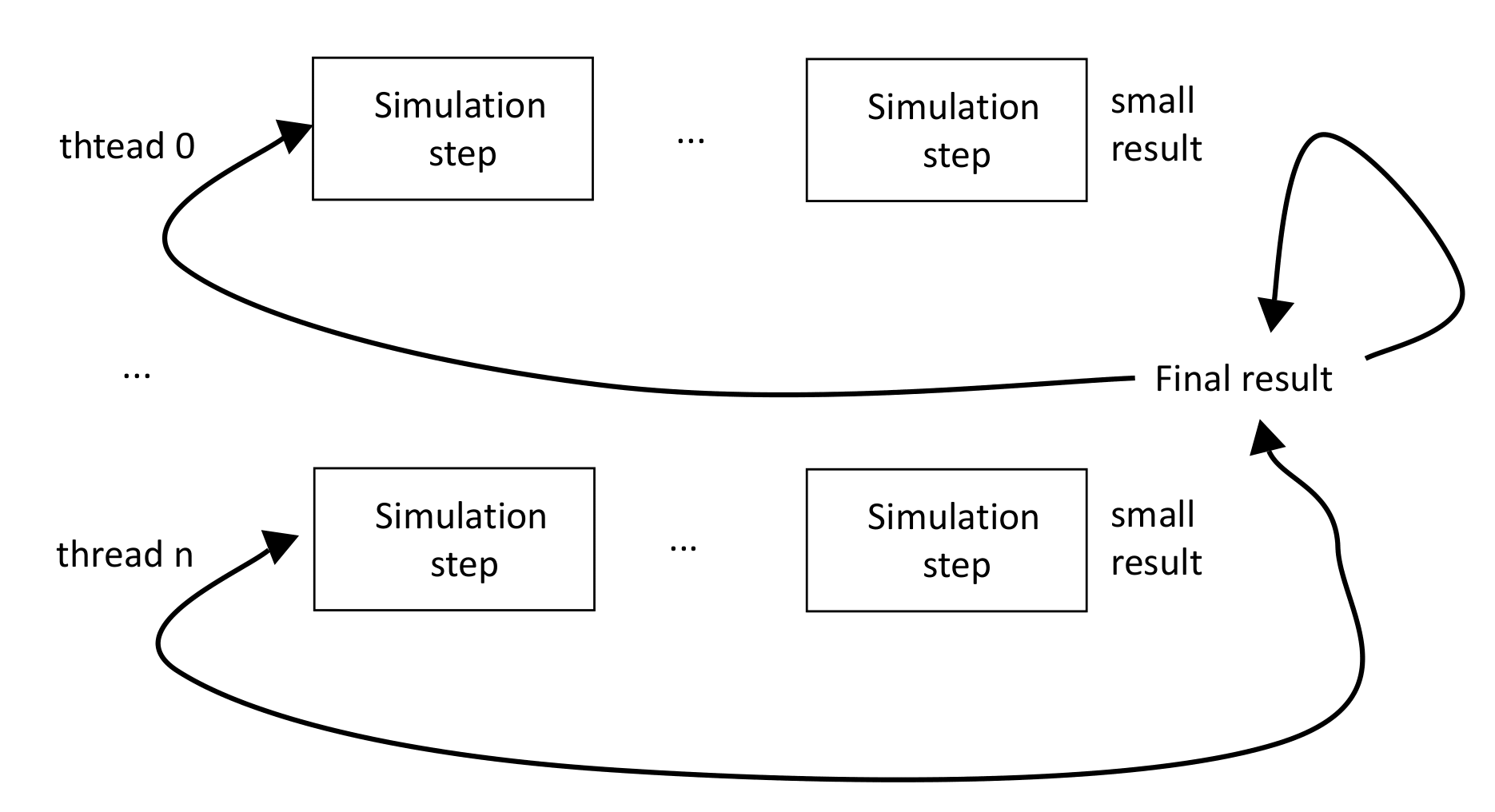}
\caption{Overall process of creating MCTS}
\label{fig:implementation}
\end{figure}

\subsection{Implementation overview}
To enhance the Monte Carlo Tree Search (MCTS) for the Da Vinci Code game, we devised a parallelized approach where each thread independently conducts a series of game simulations to generate a miniature MCTS, rooted in the current positions of both the player and the opponent. Within each thread, the game iteratively expands a child node from the root\cite{chen2023optical}, with subsequent gameplay unfolding randomly after the initial move\cite{zou2023capacity}. The outcomes of these simulated games inform the updates to the parent node, progressively building a comprehensive decision tree. This process of expansion and update is repeated across multiple threads, each contributing to constructing a segmented MCTS. These individual trees, resulting from separate threads, are then amalgamated to form the final MCTS structure, which underpins decision-making in the game. Figure \cref{fig:implementation} depicts this multi-threaded simulation process, culminating in a consolidated MCTS that guides player decisions at their turn.

\subsection{Implementation of \cpu}
To optimize decision-making efficiency, Monte Carlo Tree Search (MCTS) employs parallelism to expedite the construction of a robust decision tree through simultaneous computation. Leveraging OpenMP, the CPU-based MCTS implementation achieves parallel processing, with each thread executing game simulations to determine outcomes (win or lose). Upon completion, these individual results are aggregated into a unified MCTS, enhancing the algorithm's ability to make informed decisions within a reduced timeframe.

\subsection{Implementation of \gpu}
Echoing our previous discussion on the necessity of parallelism for effective Monte Carlo Tree Search (MCTS) execution, the GPU-based implementation utilizes CUDA to facilitate parallel computations. Similar to its CPU counterpart, each thread on the GPU processes multiple simulations, concluding with a binary outcome (win or lose). These outcomes from individual threads are then synthesized into a comprehensive MCTS, ensuring efficient decision-making by leveraging the computational prowess of GPUs. 

\section{Evaluation}
We implemented MCTS for Da Vinci code upon a 4-CPU(Pascal Titan Xp) and 12-core CPU. We measured the execution time to analyze the dominant factor of time consumption. We evaluated the performance and difference between \cpu and \gpu through measuring the number of simulation per second.

\subsection{Execution Time}
\begin{figure}
\includegraphics[width=0.95\columnwidth]{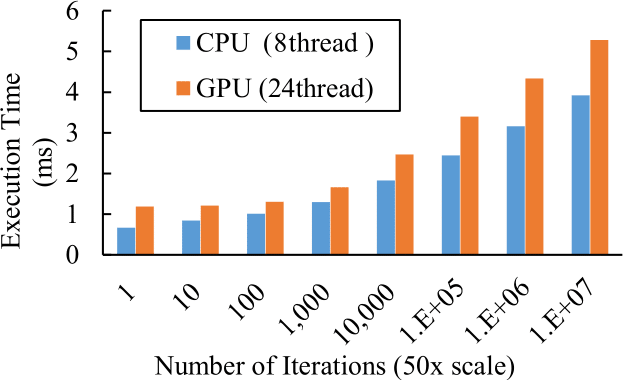}
\caption{The time-consuming according to the number of iterations}
\label{fig:time_consuming}
\end{figure}

We measured the execution time by varying the number of simulations from 1 to $10^7$ and calculated average values on 5 times execution. 
The \cref{fig:time_consuming} shows the execution time which was measured using 12 threads upon \cpu and using 32 threads upon \gpu.
We could observe that the execution time increases as the number of simulations increases~\cite{song2023deep}.
Therefore, the number of simulations can be a dominant factor of time consumption. 

The \cref{fig:time_consuming} shows that the execution is not much different when the number of simulations is small. 
The reason is when the number of simulations is small, the memory access to the large array becomes the key factor of the performance. 
However, as the number of simulations increases, the number of simulations and the execution time are proportional.
This is because when the number of simulation is large, computing becomes the key factor\cite{gao2023autonomous}. 
Therefore the performance (execution time) is proportional to the number of simulations if the amount of computation is large.

\subsection{The Number of Simulations per Second}
\subsubsection{CPU}
\begin{figure}
\includegraphics[width=0.95\columnwidth]{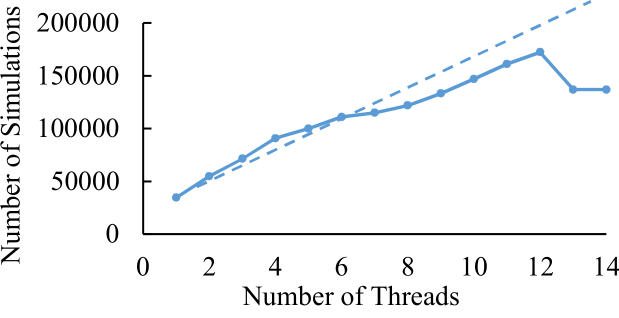}
\caption{The number of simulations per second on CPU}
\label{fig:cpu_num_simulation}
\end{figure}

The \cref{fig:cpu_num_simulation} shows the result of the number of simulations per second over increasing number of threads of CPU. 
As increasing threads, the number of simulations per second also increases linearly. 
However, you can see the number of simulations decreases if the number of threads exceeds 12. 
The reason why the number of simulations decrease is that we have only 12 core. In other word, the number of physical cores is the scalability bottleneck. 
Therefore, it can be seen that the performance increases linearly with the number of cores in the CPU. 
\subsubsection{GPU}
\begin{figure}
\includegraphics[width=0.95\columnwidth]{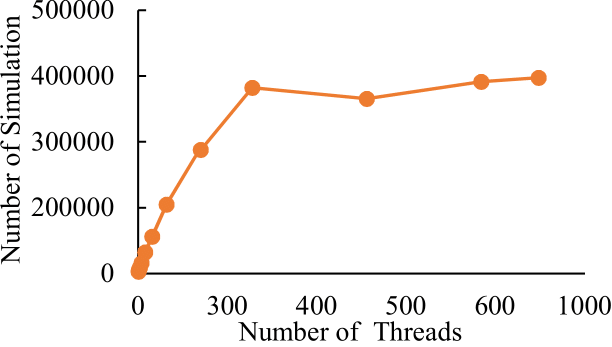}
\caption{The number of simulations per second on GPU}
\label{fig:gpu_num_simulation}
\end{figure}

The \cref{fig:gpu_num_simulation} shows the number of simulations per second over the increasing number of threads of GPU. 
Unlike the \cref{fig:cpu_num_simulation}, the \cref{fig:gpu_num_simulation} shows the non-linear fashion. 
In addition, the graph has instant performance degradation at the middle despite increasing the number of threads. 
This is because cache misses increase as the number of threads is used, and memory bottlenecks from cache misses are higher than computation performance obtained by increasing threads in GPU.

We measured the number of simulations with varying the number of warps to identify the valley mentioned in ~\cite{li2015priority_valley}. 
\begin{figure}
\includegraphics[width=0.95\columnwidth]{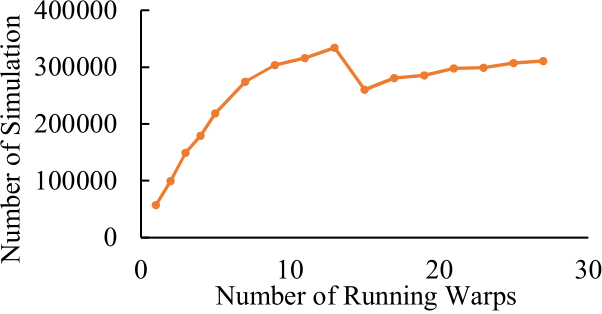}
\caption{The number of simulations according to the number of warps on GPU}
\label{fig:gpu_warp_simulation}
\end{figure}

The \cref{fig:gpu_warp_simulation} shows the experiment result that measured the number of simulations according to the number of warps. 
If the number of warps is over 12, the performance decreases like ~\cite{zhou2023semantic}. 
The memory contention makes valley, therefore the number of simulations reduces despite increasing the number of warps. 

\begin{figure}
\includegraphics[width=0.95\columnwidth]{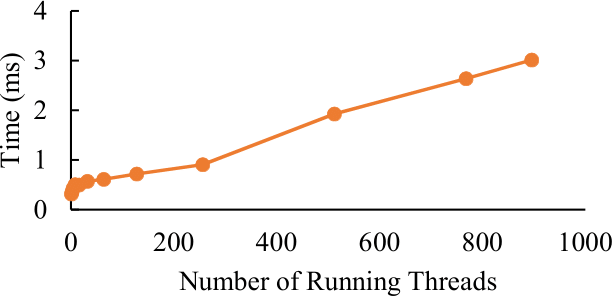}
\caption{Time of executing the application on GPU depending on thread size.}
\label{fig:gpu_thread_time}
\end{figure}

In MCTS based on GPU, the threads that are already finished should wait until the thread with the most turns ends. Because other threads need to wait for the longest thread, the MCTS based on GPU should be the weak scaling which takes the same amount of time to increase the number of threads.
However, the \cref{fig:gpu_thread_time} shows that weak scaling is not sustained. 
There are two reasons why the \cref{fig:gpu_thread_time} is not weak scaling. 
First is path divergence issue until 16 threads. 
Many divergences to guess the opponent's tile take more time to finish the computation.   
Second is the memory contention issue which performance valley as we mentioned previously more than 16 threads.
As the number of threads increases, not enough memory makes memory contention so the time takes more time.

\section{Conclusion}
We implement $\cpu$ and $\gpu$ for Da Vinci code for parallelization. To Compare $\cpu$ with $\gpu$, we measure the execution time by increasing the number of simulations and the number of simulations with growing thread. Through the result, $\cpu$ shows well enormously scaled in parallelization while $\gpu$ shows non-weakly scaled due to memory contention. 

\bibliographystyle{IEEEtran}
\bibliography{reference}

\end{document}